\pdfoutput=1

\documentclass[11pt]{article}

\usepackage[final]{acl}
\usepackage{amsmath}
\usepackage{times}
\usepackage{latexsym}
\usepackage{tabularx}
\usepackage[T1]{fontenc}

\usepackage[utf8]{inputenc}
\usepackage{soul}
\usepackage{microtype}

\usepackage{inconsolata}

\usepackage{graphicx}
\usepackage{subcaption}
\usepackage{array}

\newcommand{\name}{MorphNLI}

\title{MorphNLI: A Stepwise Approach to Natural Language Inference Using Text Morphing }

\author{Vlad-Andrei Negru$^{1}$, Robert Vacareanu$^{1,2}$, Camelia Lemnaru$^{1}$, \\ \textbf{Mihai Surdeanu$^{2}$, Rodica Potolea$^{1}$ }\\\\
$^1$Department of Computer Science, Technical University of Cluj-Napoca, Cluj-Napoca, Romania  \\
	$^2$Department of Computer Science, University of Arizona, Tucson, USA \\ 
		\{vlad.negru, camelia.lemnaru, rodica.potolea\}@cs.utcluj.ro, \\\{rvacareanu, msurdeanu\}@arizona.edu  \\
}

\newcolumntype{Y}{>{\centering\arraybackslash}X}
\newcolumntype{F}[1]{%
    >{\raggedright\arraybackslash\hspace{0pt}}p{#1}}%

\newcolumntype{Z}{>{\raggedleft\arraybackslash}X}

\begin{document}
\maketitle
\begin{abstract}
We introduce MorphNLI, a modular step-by-step approach to natural language inference (NLI). When classifying the premise-hypothesis pairs into \{\textit{entailment}, \textit{contradiction}, \textit{neutral}\}, we use a language model to generate the necessary edits to 
incrementally
transform (i.e., \textit{morph}) the premise into the hypothesis.
Then, using an off-the-shelf NLI model we track how the entailment progresses with these atomic changes, aggregating these intermediate labels into a final output.
We demonstrate the advantages of our proposed method particularly in realistic cross-domain settings, where our method always outperforms strong baselines, with improvements up to 12.6\% (relative). 
    Further, our proposed approach is explainable as the atomic edits can be used to understand the overall NLI label.

\end{abstract}

\section{Introduction}
\begin{figure*}[h]
    \centering
    \includegraphics[width=1\textwidth]{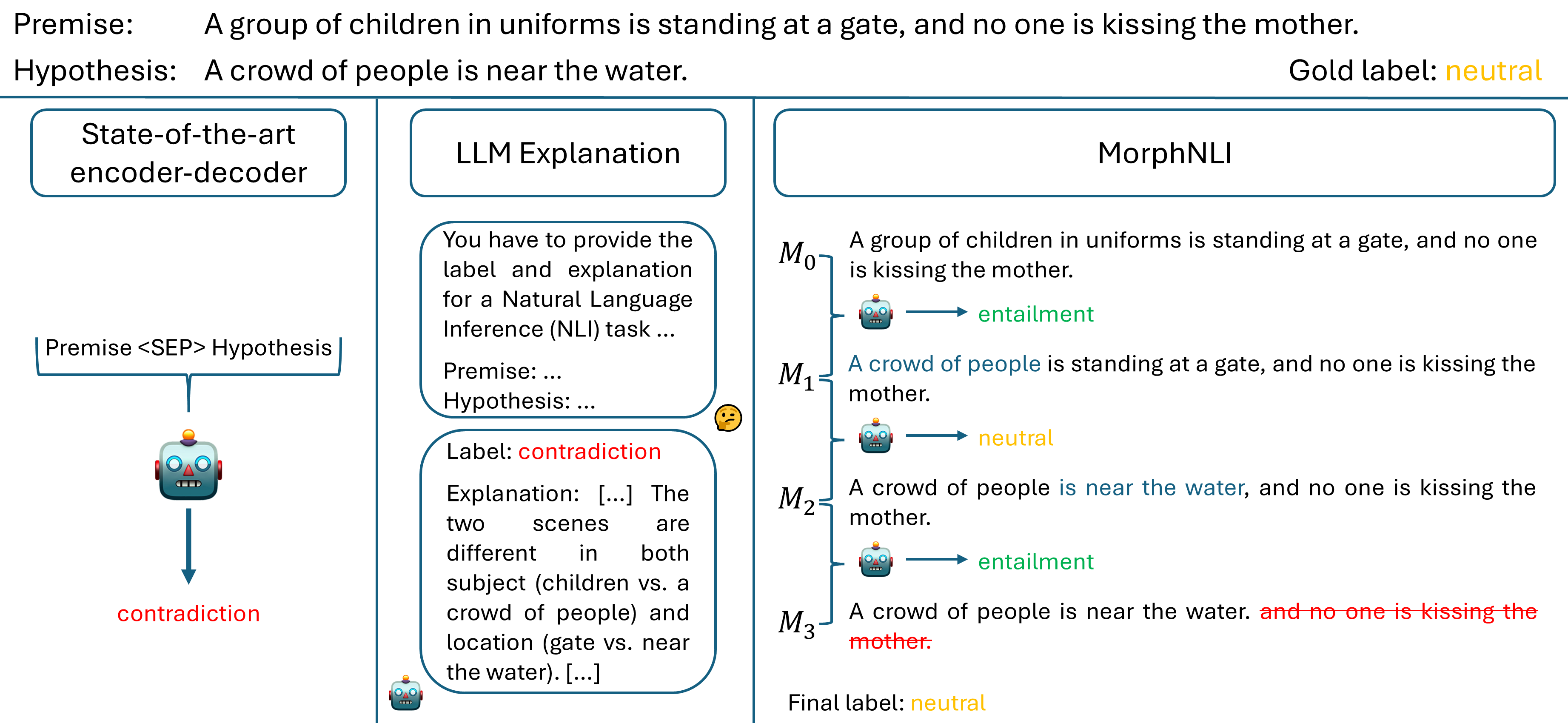}
    \vspace{-4mm}
    \caption{Natural language inference example where both a state-of-the-art encoder-decoder model -- BART (left) and a LLM -- GPT-4o (middle) predict the incorrect label. Our approach  (right) incrementally morphs the premise into the hypothesis, which 
    decomposes the inference process into several simpler steps. This allows 
    it to generate the correct label, which is also associated with an intuitive explanation that falls naturally from the morphing steps. In contrast, both the encoder-decoder model and the LLM produce the incorrect label. The LLM's explanation suggests overfitting on annotation artifacts from SNLI, which assumes coreference between participants and concepts in the two texts~\cite{jiang-marneffe-2022-investigating}.
    }
    \vspace{-4mm}
    \label{fig:example}
\end{figure*}

Natural Language Inference (NLI), i.e., the task that determines whether a text hypothesis is true, false, or undetermined given a text premise \cite{condoravdi-etal-2003-entailment, Dagan2005ThePR, bowman2015large}, is an important building block of many applications such as question answering, summarization, and dialogue systems, where understanding the logical connection between different pieces of information is essential \cite{yin-etal-2019-benchmarking, sainz-etal-2021-label, sainz-etal-2022-textual}. 
Despite the fact that NLI has received significant attention lately \cite{Raffel2019ExploringTL, Jiang2019SMARTRA, Sun2020SelfExplainingSI, Wang2021EntailmentAF}, 
several analyses have indicated that neural NLI methods fail to capture important semantic features of logic such as monotonicity, and more granular aspects like negation, universal vs. existential quantifiers, and concept modifiers~\cite{rozanova-etal-2022-decomposing,akoju-etal-2023-synthetic}. Other significant limitations of current models are caused by task artifacts that oversimplify the NLI problem~\cite{williams-etal-2018-broad,jiang-marneffe-2022-investigating}. Large Language Models (LLMs) are prone to contamination~\cite{golchin2024time,contindex}, which causes overfitting on these task artifacts (see Section \ref{sec:discussion}). LLMs also tend to ``not say what they think'' \cite{turpin2024language}, which reduces the quality and faithfulness of their explanations. 

To address the above drawbacks, we propose a {\em cautious} NLI strategy that decomposes the NLI decision into several simpler and more explainable steps. Specifically, our approach: (a) incrementally transforms the premise into the hypothesis using text morphing~\cite{huang2018text}; (b) applies an off-the-shelf NLI model on each morphing iteration; and (c) aggregates the individual NLI labels into an overall label for the given premise-hypothesis pair. We call our method \name. Figure~\ref{fig:example} provides a walk-through example of our approach, contrasted with a state-of-the-art encoder-decoder model and an LLM. The advantages of our direction are two fold. First, it performs better out of domain because its individual, smaller decisions reduce the chance of overfitting. Second, it naturally produces an explainable reasoning chain that traces the morphing transformations. 

Our approach is inspired by Natural Logic (NL)~\cite{maccartney2009extended,maccartney2014natural} but is more flexible. First, rather than relying on a formal alignment algorithm between premise and hypothesis, which continues to be a pain point in the development of NL systems~\cite{krishna2022proofver}, we use a more nimble morphing algorithm \cite{huang2018text} that is trained on synthetic data. Second, instead of using the seven NL logic operators and a relatively complex finite-state automaton to aggregate them, we rely just on the three standard NLI labels (entailment, contradiction, neutral) and 
on a straightforward, robust aggregation decision that performs well in practice: pick the first non-entailment label in the sequence of NLI decisions.

The contributions of our paper are:
{\flushleft {\bf (1)}} We introduce \name, a modular approach for NLI that combines text morphing with neural NLI. Our method does not require any additional supervision, i.e., the text morphing model is trained using synthetic data; the neural NLI engine is an off-the-shelf model.
{\flushleft {\bf (2)}} We evaluate our proposed method in multiple scenarios, including two cross-domain settings: from MNLI~\cite{williams-etal-2018-broad} to SICK~\cite{sick}, and from SICK to MNLI. Our empirical evaluation indicates that \name\ outperforms other state-of-the-art NLI models in all cross-domain experiments. Further, morphing improves the decisions of GPT-4o in the SICK dataset, further highlighting that LLMs do not capture well the semantics of logic~\cite{rozanova-etal-2022-decomposing,akoju-etal-2023-synthetic}.
{\flushleft {\bf (3)}} We perform a qualitative analysis of the explanations generated by \name, and show that they are  better than GPT-4o's on SICK, despite the fact that our model sizes are orders of magnitude smaller. However, both NLI performance and explanation quality of \name\ are worse on MNLI, which we suspect is due to the LLM's contamination with the MNLI dataset.

\begin{figure*}[th!]
    \centering
    \includegraphics[width=0.95\linewidth]{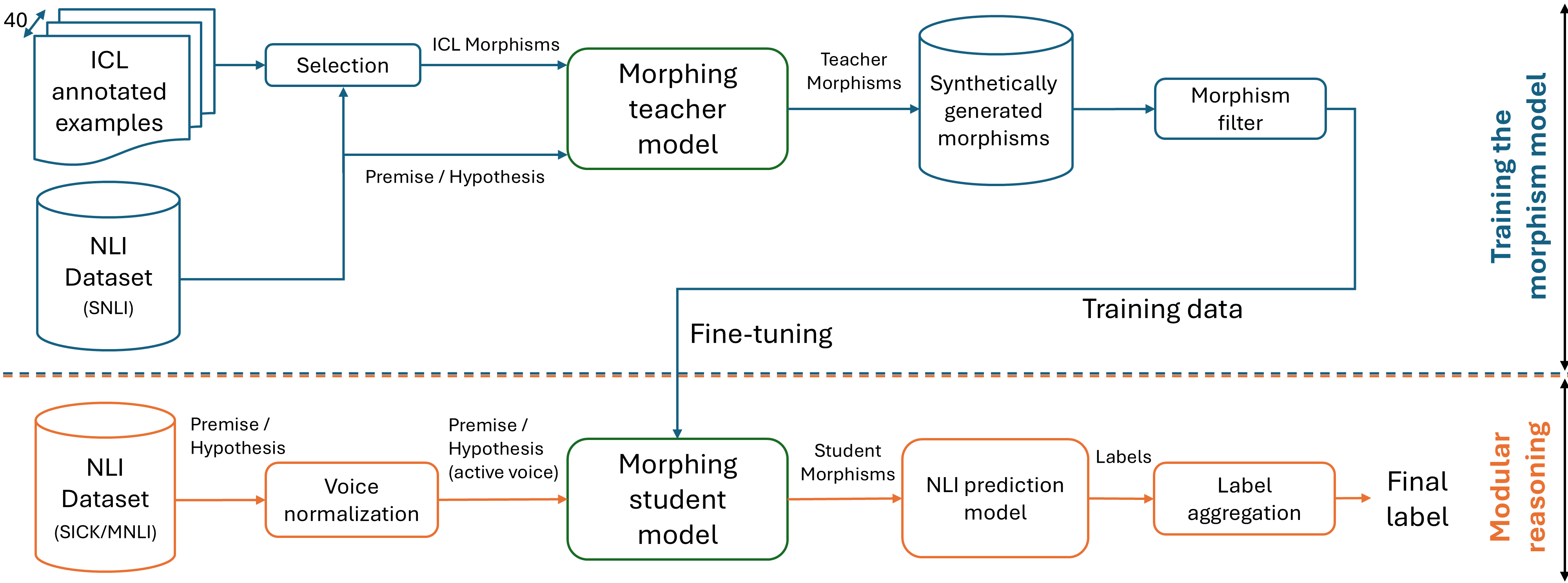}
    \vspace{-1mm}
    \caption{Training (top) and inference (bottom) for \name, including synthetic data generation for morphing. For the teacher model we use GPT-4; for the student model we use GPT-4o-mini.} 
    \label{fig:archtecture}
    \vspace{-4mm}
\end{figure*}

\section{Related work}
Our work draws inspiration from Natural Logic \cite{Lakoff1970LinguisticsAN}, which is a form of reasoning aiming to draw logic inferences by operating directly over linguistic structures. 
Over the years, this has been implemented for natural language processing in various forms \cite{MacCartney2007NaturalLF, Krishna2021ProoFVerNL, rozanova-etal-2022-decomposing, feng-etal-2022-neuro, korakakis-vlachos-2023-improving}. \citet{MacCartney2007NaturalLF} introduced one of the first computational models for natural logic, which has been subsequently extended and improved in follow up work \cite{MacCartney2008ModelingSC, MacCartney2009AnEM}. 
Natural logic can be useful beyond natural language inference, for tasks such as commonsense reasoning \cite{angeli-manning-2014-naturalli}, fact verification \cite{krishna2022proofver, Strong2024ZeroShotFV}, or polarity tracking \cite{hu-moss-2018-polarity}. 
One drawback of natural logic is that it is too strict. For example, natural logic cannot readily accommodate paraphrases or temporal reasoning. Our proposed approach relaxes the strict requirements of natural logic formalism, relying instead on text morphing \cite{huang2018text} and off-the-shelf NLI models. 

Our work is also related to explainable NLI \cite[ inter alia]{camburu2018explainablenli, Thorne2019GeneratingTE, camburu-etal-2020-make}. Importantly, in our proposed approach, the explanations are guaranteed to be faithful \cite{Kumar2020NILEN}, as they are constructed based on the atomic edits produced by the morphing model.

Tangentially, our proposed approach resembles work on modeling edit processes \cite{guu-etal-2018-generating, awasthi-etal-2019-parallel, reid-neubig-2022-learning, reid2023diffuser}. Very relevant is the work on text morphing \cite{huang2018text}, which we repurpose to generate atomic edits to transform the premise into the hypothesis. 

We also leverage off-the-shelf NLI models to produce the final label. We refer the interested reader to the survey of \citet{Storks2019RecentAI}. Specifically, we use transformer-based NLI models \cite{Vaswani2017AttentionIA, Devlin2019BERTPO, Liu2019RoBERTaAR, Lewis2019BARTDS}, typically trained on a mixture of NLI datasets \cite{sick, bowman2015large, williams-etal-2018-broad}. 








\section{Approach}

Our proposed method, \name, uses a modular step-by-step approach for natural language inference (NLI). At a high level, \name\ operates in three steps: (a) the premise is incrementally converted into the hypothesis through a sequence of small atomic edits that we call {\em morphisms} (see subsection~\ref{sec:morphisms}); (b) an NLI engine is applied to generate NLI labels for each pair of texts in the sequence of transformations; and (c) these labels are aggregated into an overall NLI label for the original premise/hypothesis pair. This is beneficial for two reasons. First, the differences between a premise and a hypothesis are gradually broken into multiple sentences, which makes the task easier for an NLI engine and less prone to overfitting. Second, the trace resulting from the atomic edits can be used as a rationale for the final label, making the method more explainable.

Figure \ref{fig:archtecture} shows the overall architecture of our pipeline. The first module presents the training of the morphism model, where we use In-Context Learning (ICL) with an LLM as a teacher model to generate a synthetic dataset labeled with morphisms. After a filtering step, we use this dataset for fine-tuning a student model for morphism generation. At inference time, we use the student model for generating the morphisms and an NLI prediction model for generating labels. The labels are then aggregated into one final prediction. 
We detail all these components below. 

\subsection{The text morphing task}
\label{sec:morphisms}

Before describing these components,  
we define the morphism generation task, similar in nature with the work of \citet{huang2018text}.
Formally, this task is the process of changing one initial sentence (i.e., premise) into a destination sentence (i.e., hypothesis) through a series of 
morphing operations. These operations are similar to the steps used in computing the Levenshtein distance: 

\begin{enumerate}
    \vspace{-2mm}
    \item Replace - (\textit{replace}, <old\_text>, <new\_text>)
    \vspace{-3mm}
    \item Remove - (\textit{remove}, <text>)
    \vspace{-3mm}
    \item Insert - (\textit{insert}, <text>)
    \vspace{-2mm}
\end{enumerate}

There are three important differences between our morphing and Levenshtein distance. First, our morphing operations 
operate at word/phrase granularity rather than characters. 
Second, our transformations are encouraged to preserve the syntactic structure of the source sentence (see subsection~\ref{subsec:training}).
Third, morphisms are generated using an LLM rather than an edit distance algorithm.

Morphing a premise into the corresponding hypothesis results in a finite sequence  $\mathbf{M}$ of sentences (morphisms), where each sentence $\mathrm{M}_i$ is the result of applying a morph operation on the previous sentence $\mathrm{M}_{i-1}$. The first sentence in this sequence is the premise and the last is the hypothesis.

\subsection{Training the morphism model}
\label{subsec:training}

One of our key contributions is training a morphism generation model with minimal supervision. 
The only supervision we require is: (a) a dataset of premise/hypothesis pairs with the associated NLI labels; and (b) a small pool of sentence pairs annotated with morphisms. 
To generate synthetic training data for morphisms we use an LLM with ICL (the teacher model). This LLM is coupled with a deterministic filter that increases the quality of the generated data. Using this data, we fine-tune a smaller LLM (the student model) to generate morphisms during inference.


\paragraph{
Morphing teacher model and ICL selection} \mbox{} \\
Given the complexity of the task and the nonexistence of a dataset labeled with morphisms, we steer the design of our method towards ICL. 
Our ICL pool contains 40 pairs of premises and hypotheses, humanly annotated with intermediate sentences and corresponding morph operations. When generating the morphisms for a pair of sentences, we select the 12 closest examples from the pool of 40 to be used in the prompt. These examples are selected based on the cosine similarity with the input premise and hypothesis, computed on the embeddings generated by a Sentence-BERT \cite{reimers-2019-sentence-bert}.

The generation of the morphisms is driven by a Chain-of-Thought \cite{Chain-of-Thought} prompt, where we ask the teacher model to output the morph operations before generating each intermediate sentence. The input prompt also contains formal rules for the morphing task, encouraging the LLM to preserve the syntactic structure of the source sentence, and forcing a strict order for the morph operations: first apply \textit{replace} operations, then \textit{remove} operations, and lastly \textit{insert} operations. We empirically found that enforcing the operations in this order improves the quality of the overall results. The complete prompt and examples of generated training morphisms are included in the appendix.

\paragraph{Morphism filter} \mbox{} \\
The synthetically annotated morphisms undergo a series of filtering steps for ensuring their quality. First, for obvious reasons, we filter out the examples where no intermediate sentences were generated (we called these examples {\em lazy morphisms}).

\begin{figure}[hp]
    \centering
    \includegraphics[width=\linewidth]{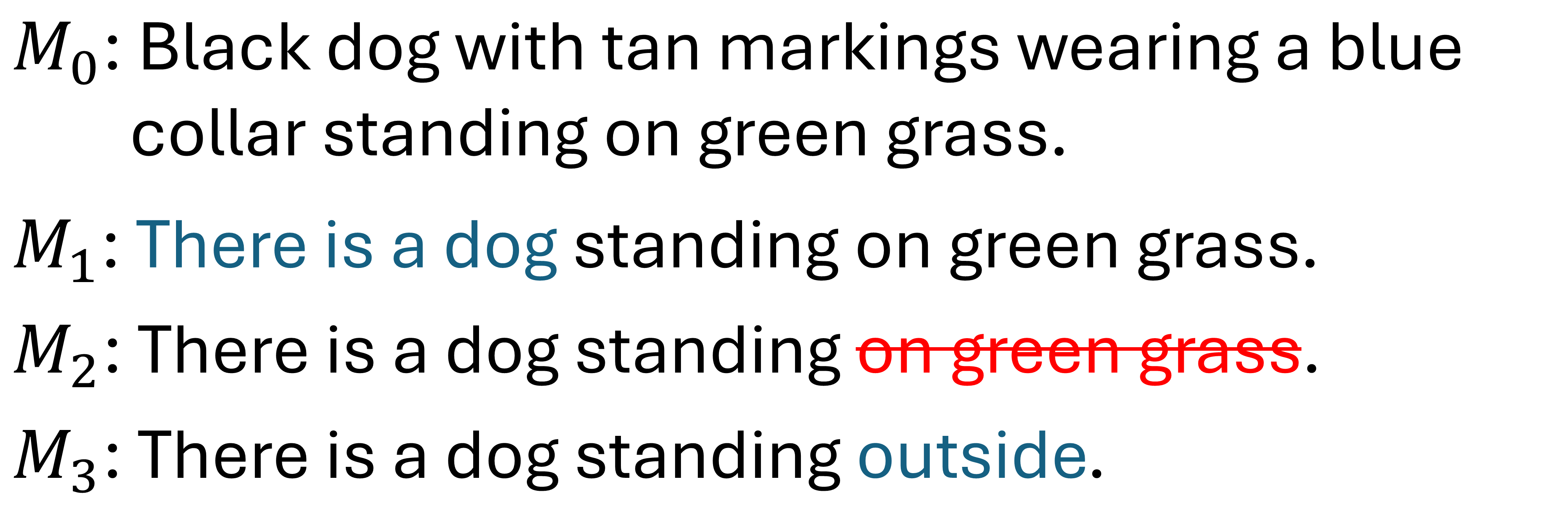}
    \caption{Example of a short morphism for sentence $M_2$. The information about the context of the action (``on green grass'') is lost when $M_2$ is generated. A similar context is then added in $M_3$ (``outside''), yielding a faulty neutral prediction because the connection ``on green grass'' $\rightarrow$ ``outside'' is lost.} 
    \label{fig:short_morphism}
    \vspace{-5mm}
\end{figure}

Second, we consider only the examples with intermediate sentences that are longer than either the premise or the hypothesis. We call the phenomenon where some intermediate sentences are too short {\em short morphisms}. This phenomenon may bring faulty reasoning processes, as some intermediate sentences may be formed by removing word groups from the initial sentence that may be necessary for future downstream NLI steps. Figure~\ref{fig:short_morphism} shows an example of this situation.
In order to limit these cases, we removed all short morphisms from the generated data. 

Last but not least, we keep only examples where the overall predicted NLI label is identical to the gold label for the given premise/hypothesis pair. Our hypothesis is that morphisms that yield the correct overall label are more likely to be correct. An initial investigation of the generated data validated our hypothesis. 
To generate individual NLI labels, i.e., between $M_{i-1}$ and $M_i$, we used a BART-large NLI classifier fine-tuned on SNLI, MNLI and FEVER; we aggregated these labels using the aggregation function described below.

\paragraph{Morphing student model} \mbox{} \\
Using the remaining synthetic data, we fine-tune a smaller LLM as the morphism student model. We used GPT-4o-mini.

\subsection{Modular reasoning using morphisms}

\begin{figure}[hp]
    \centering
    \includegraphics[width=1\linewidth]{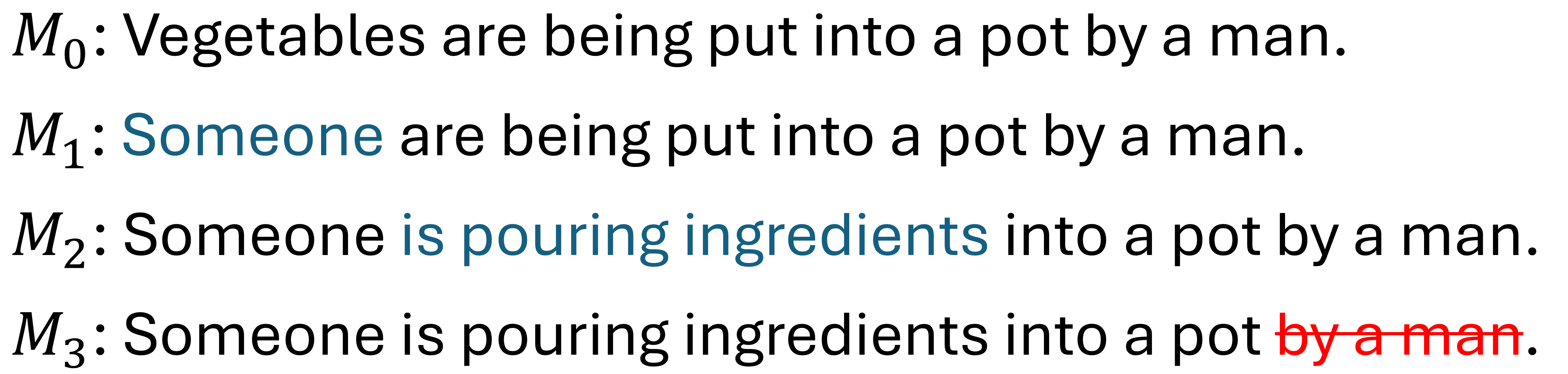}
    \caption{Example of morphisms with no voice correction. Due to the difficulties caused by the change from passive to active voice between premise and hypothesis, the morphing model  ``hallucinates'' inner sentences.}
    \label{fig:voice_correction}
    \vspace{-5mm}
\end{figure}

During inference, \name\ operates in 4 steps:
{\flushleft {\bf (1) Voice normalization (VN):}} We observed that the sequential nature of morphing operations proves to be too rigid when there is a change of voice between the premise and hypothesis, as Figure~\ref{fig:voice_correction} shows. To address this, we normalize the premise and the hypothesis to active voice using a smaller language model.

{\flushleft {\bf (2) Morphing:}} We use the above morphism student model to generate the transformations between the premise and hypothesis.

{\flushleft {\bf (3) Generating individual NLI decisions:}} We use an existing NLI classifier to generate the individual NLI labels between every ($M_{i-1}$, $M_i$) pair of sentences capturing a morphing transformation (see Figure~\ref{fig:example} for an example).

{\flushleft {\bf (4) Aggregating NLI decisions:}} An aggregation function is then used to combine the sequence of NLI labels into an overall label for the given premise/hypothesis pair. To this end, we use a simple heuristic: if all individual NLI labels are {\em entailment}, then the overall label is  {\em entailment}; otherwise the overall label is set to be the first (left-most) individual label that is not {\em entailment}. For example, in Figure~\ref{fig:example} the first non-entailment label is {\em neutral}, which becomes the overall prediction for the example in the figure. In initial experiments, we experimented with aggregating labels using the Natural Logic fine state automaton~\cite{maccartney2014natural,krishna2022proofver}, but have observed that this more formal automaton does not translate well to our more flexible setting. In contrast, our heuristic performed better and is efficient, as it does not require substantial additional processing overhead.

\section{Experimental results}
\subsection{Datasets used}

We evaluate the NLI performance of \name\ using two datasets: Multi-Genre Natural Language Inference (MNLI) \cite{williams-etal-2018-broad} and Sentences Involving Compositional Knowledge (SICK)~\cite{sick}. MNLI covers 10 genres of written and spoken English and contains fairly complex natural language. SICK contains artificially-generated premise/hypothesis pairs, which were created using a formal set of logic rules that follow syntactic and lexical transformations. As such, SICK exhibits different challenges from MNLI, assessing the ability of NLI models to comprehend complex logic and compositional structures. Considering these differences, these two datasets are a good selection for both in-domain (ID) and out-of-domain (OOD) evaluations. That is, in addition of training and testing in each dataset, we evaluate \name\ when the underlying NLI engine is trained on the other dataset.

We did not use the Stanford Natural Language Inference (SNLI) corpus \cite{bowman2015large} for the NLI evaluations because, as some of its original authors noticed, it ``is not sufficiently demanding to serve as an effective benchmark''~\cite{williams-etal-2018-broad}. SNLI ignores important phenomena such as temporal reasoning, compositionality of logic, and they make simplifying coreference assumptions, i.e., that the participants and concepts mentioned in the premise and hypothesis are the same~\cite{williams-etal-2018-broad,jiang-marneffe-2022-investigating}. 

However, to minimize any potential overfitting, we fine-tune the morphing engine using premise/hypothesis pairs from SNLI (see next subsection). Thus, our morphing component can be seen as always being evaluated out-of-domain.

\subsection{Experimental settings}


The LLMs used for the teacher and student morphing models are both from the GPT-4 family. Details on the models' identifiers are present in Appendix \ref{sec:models_used}, together with experiments using another LLM from the Llama family. The synthetic data set that contains morphisms is generated from the SNLI validation dataset (\textasciitilde10,000 premise/hypothesis pairs) using GPT-4-turbo. After the filtering step (see Section \ref{subsec:training}), we are left with 3,027 pairs labeled with morphisms for fine-tuning. These are split into two sets: one for training (2,127 examples) and one for validation (900 examples). The remaining filtered-out examples are later used to compare the fine-tuning approach with simple ICL for morphing. Our 
preliminary experiments indicated that fine-tuning outperforms ICL, both in terms of overall performance and model efficiency. For this reason, all experiments described later in this section use a morphing model fine-tuned on the above training data. Moreover, the fine-tuned model proves to be more expressive, with a much lower rate of lazy morphisms (1,575 compared to 4,375 in the case of ICL), having 
a slight increase in the number of short morphisms (674 compared to 557 in the case of ICL).


For the individual NLI decisions we experimented with state-of-the-art NLI prediction models from two different families: encoder-decoder using BART and encoder-only using RoBERTa (large versions).\footnote{The sources of the models are presented in the appendix.} 






\subsection{Results}
\label{sec:results}

Table \ref{tab:SICK} shows the accuracy of MorphNLI on the SICK test dataset. We report the results with and without voice normalization, with two different NLI engines (RoBERTa and BART), which are trained both ID and OOD. We compare the performance of our approach to the same NLI model applied directly to the original premise/hypothesis pair, i.e., without morphing (referred to as ``vanilla'').
We draw several observations from this table. First, all models perform better ID than OOD, which indicates a certain degree of overfitting. 
Second, \name\ shows a slight drop in ID performance, which we attribute to the fact that the NLI models were not trained on incremental transformations (see the next subsection for a more detailed analysis). Most importantly, \name\ outperforms the ``vanilla'' NLI model in all four OOD configurations, with an improvement as large as 1.74\% for BART with voice normalization. This is an encouraging result, as it validates our hypothesis: that modular NLI improves domain transfer.

\begin{table}

\begin{center}
\resizebox{1\linewidth
}{!}{
\begin{tabularx}{1.2\linewidth}{ l *{2}{Z} }
\hline
\textbf{SICK} & \textbf{ID}& \textbf{OOD} \\
\hline
RoBERTa Vanilla  & 90.64 & 56.62 \\
RoBERTa Vanilla (+VN) & \textbf{90.91} & 56.52 \\
RoBERTa Morphism  & 88.14 & 57.68\\
RoBERTa Morphism (+VN) & 88.32 & \textbf{57.94}\\
\hline
BART Vanilla  & 89.85 & 59.29   \\
BART Vanilla (+VN) & \textbf{90.07} & 58.64   \\
BART Morphism  &  87.38 & 59.64   \\
BART Morphism (+VN) &  88.59 & \textbf{60.38}   \\
\hline
\end{tabularx}
}
\end{center}
\vspace{-2mm}
\caption[cap]{\name\ accuracy on the SICK dataset using two NLI engines: RoBERTa and BART. We compare our results against the two ``vanilla'' NLI models, i.e., without using text morphing. 
VN indicates voice normalization. For OOD, we use the RoBERTa models trained on MNLI, and BART models trained on SNLI, MNLI and FEVER.\footnotemark}
\label{tab:SICK}
\vspace{-3mm}
\end{table}

\footnotetext{We empirically observed that this model outperforms an MNLI-only trained BART.}

Table \ref{tab:MNLI} shows the same behaviour for the MNLI test dataset. Here the OOD enhancements are more considerable. For example, we observe an increase of 5.29\% for RoBERTa and 5.92\% for BART, both in settings with no voice normalization. While the voice normalization proved beneficial for the SICK dataset, for all the scenarios tested, for MNLI we see a decline in accuracy when applying it (see the next subsection for a more detailed discussion). 

\begin{table}

\begin{center}
\resizebox{1\linewidth
}{!}{
\begin{tabularx}{1.2\linewidth}{ l *{2}{Z} }
\hline
\textbf{MNLI} & \textbf{ID}& \textbf{OOD} \\
\hline
RoBERTa Vanilla  & \textbf{89.91} & 53.00 \\
RoBERTa Vanilla (+VN) & 88.50 & 52.77 \\
RoBERTa Morphism  & 85.01 & \textbf{58.29} \\
RoBERTa Morphism (+VN) & 83.32 & 56.73 \\
\hline
BART Vanilla  & \textbf{88.24} & 46.86   \\
BART Vanilla (+VN) & 86.48 & 45.50  \\
BART Morphism  &  82.00 & \textbf{52.78} \\
BART Morphism (+VN) &  80.16 & 51.12   \\
\hline
\end{tabularx}
}
\end{center}
\vspace{-2mm}
\caption{\name\ accuracy on the MNLI dataset, under the same settings as Table \ref{tab:SICK}. For OOD, we use models trained on SICK.}
\label{tab:MNLI}
\vspace{-3mm}
\end{table}


To understand if our approach is compatible with LLMs, we evaluate the performance of our pipeline in another setting, 
in which we use two LLMs (GPT-4o and GPT-4o-mini) as the NLI engines. These results are presented in Table \ref{tab:gpt}. 
Despite their massive size and their extensive training, these LLMs still benefit from morphing on the SICK dataset. This underlines previous observations that LLMs do not capture well the semantics of logic, which is a key focus in SICK~\cite{rozanova-etal-2022-decomposing, akoju-etal-2023-synthetic}. However, we do not observe a similar improvement on MNLI. One potential explanation for such an effect is the potential contamination of these LLMs with the MNLI dataset (see next subsection for a longer discussion).

\begin{table}[htbp]

\begin{center}
\resizebox{1\linewidth
}{!}{
\begin{tabularx}{1.2\linewidth}{l*{2}{Z}}
\hline
\textbf{Model} & \textbf{SICK}& \textbf{MNLI} \\
\hline
GPT-4o Vanilla  & 60.38 & \textbf{83.58} \\
GPT-4o Morphism & \textbf{61.05} & 73.55 \\
\hline
GPT-4o mini Vanilla & 62.25 & \textbf{79.68}   \\
GPT-4o mini Morphism &  \textbf{62.86} & 73.13   \\
\hline
\end{tabularx}
}
\end{center}
\vspace{-2mm}
\caption{GPT-4o and GPT-4o-mini NLI accuracies, with and without text morphing.} 
\label{tab:gpt}
\vspace{-6mm}
\end{table}

\subsection{Discussion}
\label{sec:discussion}

To further understand \name's behavior we answer below three important research questions.


\subsubsection{What is the quality of \name's explanations?}
\label{ssec:manual}

To get a better understanding of how the explanations generated via our modular approach compare to those generated by LLMs (GPT-4o and Llama 3.1 8B), we performed a manual evaluation on a random sample from both MNLI and SICK datasets. We selected 20 instances from each development set, distributed as follows: 5 instances where the NLI model's predictions are correct both with and without morphing, 5 where both predictions are incorrect, 5 where morphing improves the NLI prediction, and 5 where it worsens the prediction. The NLI model used here was RoBERTa, fine-tuned in-domain on each dataset. Four human evaluators awarded a score between 0 and 2, as follows: 2 indicates the explanation is correct; 1 indicates the explanation is partially correct, i.e.,  it contains correct elements, but it misses required information or includes extraneous elements; and 0 indicates the explanation is completely incorrect.

We computed Cohen’s Kappa inter-annotator agreement across all six pairs of evaluators from the four annotators. For the MNLI dataset, we calculated an average Kappa agreement of 34\%, which indicates fair agreement. Considering the complexity of the task, i.e., evaluators had to evaluate both the correctness of each morphism and the NLI label produced at each step, we consider this a respectable result. Even more encouragingly, the maximum agreement between two annotators was 57\%, which falls on the high end of moderate agreement, touching on substantial. This suggests that the agreement can be improved with more training. Similarly, on the SICK dataset, the average agreement is 67\% (substantial agreement), with a maximum of 91\% (almost perfect). These higher scores highlight that despite the complexity of the task, annotators were trained to perform high-quality annotations.

We evaluated: (i) the overall explanation quality with our modular approach; (ii) the quality of the GPT-4o reasoning process and (iii) the quality of the morphisms generated via our approach (we asked the evaluators to discard the NLI label, and reason based on the morphisms alone). Table \ref{tab:manual} presents the percentage scores average from the four annotators. 

\begin{table}[htbp]
\begin{center}
\resizebox{1\linewidth
}{!}{
\begin{tabularx}{1.2\linewidth}{l*{2}{Z}}
\hline
 \textbf{Model} & \textbf{SICK} & \textbf{MNLI} \\
\hline
\name\ explanations & \textbf{70.63}  & 70.00  \\
GPT-4o explanations & 62.50& \textbf {92.50}  \\
Llama 3.1 8B explanations & 47.50 & 56.67 \\
\hline
Morphism only & 95.63  & 82.50    \\
\hline
\end{tabularx}
}
\end{center}
\vspace{-3mm}
\caption{Average percentage scores for the quality of the explanations produced via morphing, compared with the GPT-4o and Llama 3.1 8B explanations. We also assessed the quality of the morphing process alone -- i.e., whether a human evaluator could infer the correct NLI label from the morphisms.}
\label{tab:manual}
\vspace{-3mm}
\end{table}

Our approach delivers uniform explanation quality across the two dataset samples (MNLI and SICK) and the overall quality is considerably larger than that of Llama 3.1 8B, despite the latter model's much larger size.
The quality of the GPT-4o explanations is significantly better for MNLI than for SICK, where the explanations via morphisms are superior. An interesting phenomenon reported by the annotators was related to the potential overfitting of the GPT-4o reasoning. In 6 out of the 20 examples sampled from SICK, the model incorrectly assumed that premise and hypothesis refer to the same situation, i.e., the participants and the concepts mentioned are the same between premise and hypothesis. See Figure~\ref{fig:GPT_overfit_example} for a simple example. This task artifact exists in the SNLI dataset \cite{bowman2015large}, where both premise and hypothesis are annotated given a single image \cite{jiang-marneffe-2022-investigating}, but it has been removed in more recent datasets such as SICK and MNLI. However, since SNLI has been publicly available for approximately a decade it is likely that it ``leaked'' into the GPT-4o training data, which learned this task artifact. As a consequence, the LLM mistakingly labeled these pairs as contradiction and produced completely incorrect explanations.
For MNLI, this phenomenon was not as strong, which is likely due to another contamination: GPT models are known to have been contaminated with the MNLI development partition~\cite{contindex}. The large difference between MNLI and SICK GPT-4o explanation scores supports this hypothesis.

\begin{figure}[htbp]
\vspace{-2mm}
\begin{center}
\resizebox{1\linewidth
}{!}{
\begin{tabular}{p{8.5cm}}
\hline
\textbf{P:} \textit{\underline{A dog}} has its head inside a red and green gift bag \\ 
\textbf{H:} \textit{\underline{The red and brown dog}} is not digging its nose inside a holiday gift bag \\ \\
\vspace{-5mm}
\textbf{Reasoning process:}
The premise states that "{ A dog} has its head inside a red and green gift bag," which \ul{ clearly indicates that the dog is interacting with the gift bag by having its head inside it.} The hypothesis, on the other hand, claims that "The red and brown dog is not digging its nose inside a holiday gift bag." This directly contradicts the premise because  \ul{it asserts that the dog is not interacting with the gift bag in the manner described.} Therefore, the hypothesis is false given the premise, \ul{leading to a label of contradiction.}\\
\hline
\end{tabular}
}
\end{center}
\vspace{-3mm}
\caption{Misbehavior 
of GPT-4o related to the artifacts from SNLI. 
The underlined font highlights the explanation fragments that are not correct with respect to the italic text in premise/hypothesis. Here the model incorrectly assumes that the dog in the premise being the same as the dog in the hypothesis. }
\label{fig:GPT_overfit_example}
\vspace{-3mm}
\end{figure}

\begin{figure*}[!h]
    \begin{subfigure}[b]{0.48\textwidth}
        \centering
        \includegraphics[width=1\linewidth]{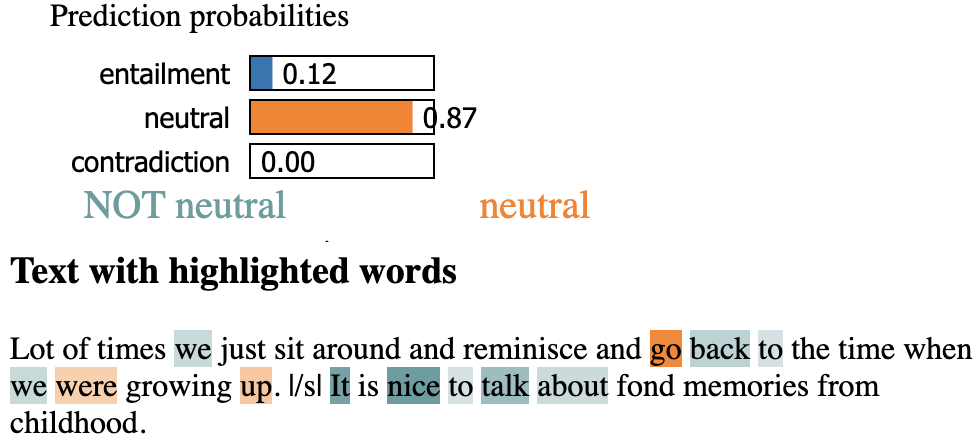}
        \subcaption{LIME explanation, no morphing}
        \label{fig:lime-1}
    \end{subfigure}
    \hfill
    \begin{subfigure}[b]{0.48\textwidth}
        \centering
        \includegraphics[width=1\linewidth]{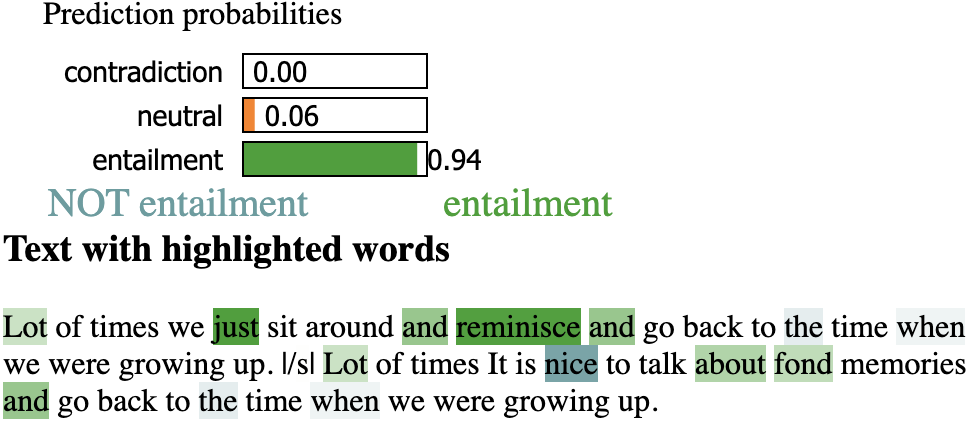}
        \subcaption{LIME explanation,  morphing step 1}
        \label{fig:lime-2}
    \end{subfigure}
    \\
    \begin{subfigure}[b]{0.48\textwidth}
        \centering
        \includegraphics[width=1\linewidth]{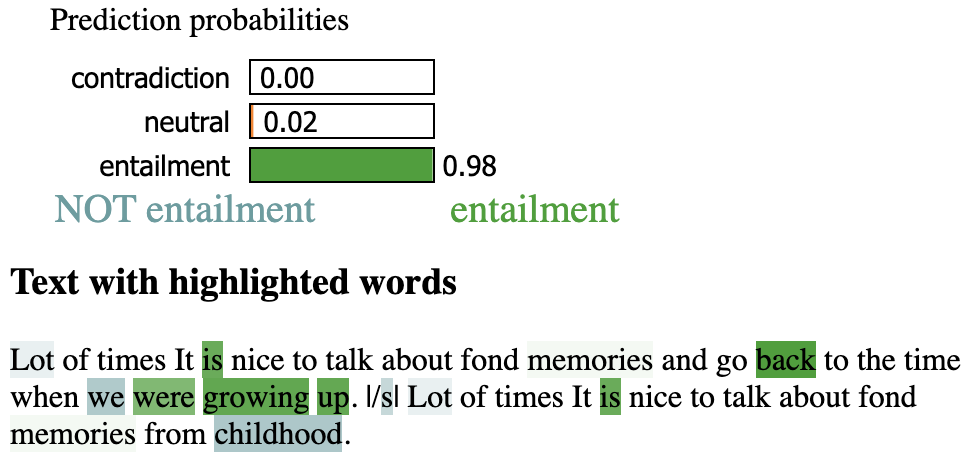}
        \subcaption{LIME explanation,  morphing step 2}
        \label{fig:lime-3}
    \end{subfigure}
    \hfill
    \begin{subfigure}[b]{0.48\textwidth}
        \centering
        \includegraphics[width=1\linewidth]{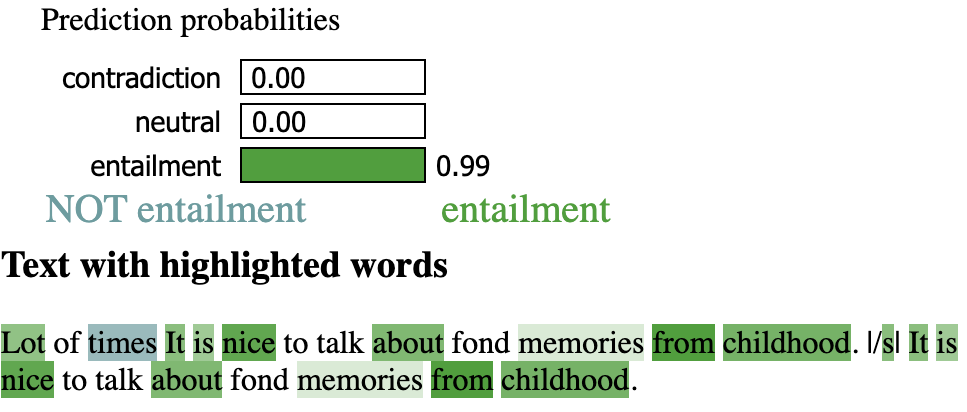}
        \subcaption{LIME explanation,  morphing step 3}
        \label{fig:lime-4}
    \end{subfigure}
    \caption{LIME analysis of the predictions of the in-domain RoBERTa NLI model for a premise-hypothesis pair from MNLI, without morphing (a) and in subsequent morphing steps (b--d). In (b) the operation is (\textit{replace}, ``we just sit around and reminisce'', ``It is nice to talk about fond memories''); in (c) the operation is (\textit{replace}, ``go back to the time when we were growing up'', ``from childhood''); in (d) the operation is (\textit{remove}, ``Lot of times'').}
    \label{fig:lime}
    \vspace{-5mm}
\end{figure*}

A second observation from this analysis is that the off-the-shelf NLI models likely exhibit some degree of overfitting as well. Specifically, they have been trained on the original premise/hypothesis pairs and underperform on our (simpler) incremental inference steps. For instance, the NLI model fails to correctly interpret semantic information in short textual snippets (e.g., understanding that "grazing a field" implies the field has "grass," or that "sandy land" implies "desert").
This explains the large difference between the morphism-only explanations (which do not use an NLI model) and the full \name\ system. In contrast, GPT-4o ---being a significantly larger model--- is able to grasp these semantic aspects. However, its explanations are sometimes long, convoluted, and repetitious, whereas the explanations provided via morphing are concise and straightforward.

\subsubsection{What are \name's common errors?}
\label{ssec:error_analysis}
To identify where most errors occur within the \name\ pipeline, we randomly sampled 20 errors from each of the two development sets (SICK/MNLI) and manually analyzed these examples. We discovered that: 45\% (SICK) and 50\% (MNLI) errors were caused by the NLI model (in-domain RoBERTa).
As indicated above, this is likely a form of overfitting due to the NLI model's original training data, which did not contain text pairs similar to our incremental transformations. This suggests that valuable future work would be to fine-tune an NLI model that is morphing aware.
The second most common error type were faulty morphisms: 45\% (SICK) and 20\% (MNLI).
This observation indicates that our morphing would probably benefit from more fine-tuning. Lastly,
5\% (SICK) and 30\% (MNLI) were a result of poor voice normalization. MNLI, which contains longer and more complex statements, potentially with several predicates, suffers from this problem more. This suggests that identifying first which verb is the sentence's main predicate might improve voice normalization.
All in all, this analysis indicates that \name's errors are caused by issues of local components, which can be potentially addressed, and are not a limitation of the overall direction.

\subsubsection{Are \name's decisions more interpretable?}
\label{ssec:lime}
To gain a better insight on how morphing improves the prediction process of the (independent) NLI model, we conducted several analyses using LIME \cite{ribeiro2016whyitrustyou} on examples from both SICK and MNLI, where we compare a ``vanilla'' NLI model (in-domain RoBERTa) with \name\ using the same NLI engine.
As anticipated, providing the NLI model with incremental changes helps it focus on more semantically relevant words. For example, Figure~\ref{fig:lime-1} shows that, without morphing, the inference model mistakenly predicts the pair as being \textit{neutral} and the focus is on words without strong relation to the sentence pair's meaning (``go,'' ``it,'' ``were,'' etc.); with morphing (Figures~\ref{fig:lime-2} --  \ref{fig:lime-4}), in each subsequent step, the model correctly and more confidently identifies all three transitions as \textit{entailment}, and focuses on more semantically relevant words (``reminisce'' and ``fond;'' ``growing up'' and ``childhood'').\footnote{Although the overall semantics are still not perfect: while the first three phrases are associated with the {\em entailment} label, ``childhood'' is associated with {\em non-entailment}.}

\subsubsection{Lexical sensitivity between the premises and hypotheses}
\label{ssec:lexical_sensitivity}

During our experiments we noticed an inverse correlation between the quality of the morphing process and the syntactic/lexical differences between the premises and hypotheses. For example, where the two share little to no lexical similarities (i.e., often found in MNLI), the morphing operations tend to consider larger textual groups, affecting the performance. On the other hand, the more similar the premise and hypothesis are, the more the morphing operations follow clear logical steps. To a large extent, this carries over to examples with larger syntactic/lexical differences. However, too many lexical differences between the premise and hypothesis hurt multiple NLI techniques, and MorphNLI is no exception. Nevertheless, our approach is less sensitive to lexical differences out-of-domain, indicating a lower degree of overfitting. This phenomenon is further detailed in the Appendix \ref{sec:appendix_lexical}.

\subsubsection{Importance of the filtering stage}

As described in section \ref{subsec:training}, we included filters to remove low-quality data, i.e., examples with no inner sentences (lazy morphisms) and examples with inner sentences that are shorter than both the premise and hypothesis (short morphisms). By filtering these examples before fine-tuning, we significantly reduce the number of short morphisms at inference time, while maintaining a low number of lazy morphisms. In our experiments, for a sample of roughly 5,500 examples during inference, the initial morphing mechanism predicted 26\% lazy and 32\% short. After introducing our filtering mechanism, the percentage of lazy morphisms had a slight increase to 28\%, while the percentage of short morphisms dropped considerably to 12\%.

\label{ssec:filtering}

\section{Conclusions}
In this paper, we proposed \name\ -- a modular step-by-step approach for natural language inference.
Our method uses a language model to generate atomic edits that progressively transform (i.e., morph) the premise into the hypothesis. We then track how these atomic edits impact the entailment between successive sentences, aggregating these intermediate labels into a final answer (see Figure~\ref{fig:example}). We hypothesized that typical NLI models can better handle examples where the two sentences are lexically close (i.e., they differ only by an atomic edit)
Our results confirm that our proposed approach is more robust, outperforming traditional NLI models in all cross-domain settings investigated.
Furthermore, our proposed method is explainable. The sequence of intermediate edits together with their corresponding individual NLI labels can be used to explain the overall prediction.

\section*{Limitations}

Our work focuses on the task of Natural Language Inference. Although the text morphing process proves to be beneficial in the context of logical reasoning, its applicability in other reasoning tasks is still to be tested. Moreover, we cannot offer an assurance on the level of generalizability of our method. For our experiments, we designed the morphism generation as a general task, as the fine-tuning data is constructed from a different domain than the testing data (SNLI vs. SICK/MNLI). However, we do not know if this generalization is constrained on data specific to the NLI task. 

In the development of our solution, for the morphism generation task, we have experimented mostly with LLMs from the GPT family. We are unsure if our pipeline may have different behavior for other proprietary LLMs or for much larger LLMs, as we are using a fine-tuned version of GPT-4o-mini. Also, being a closed source model, we do not know if the LLM was previously trained towards this objective of text morphing. Further, it is hard to accurately predict the level of contamination of the model with the test datasets, and what influence it has on the morphism generation. 

The morphing process is evaluated only on English. We have no assurance that the same techniques could apply on other languages for multilingual models or if it follows only the particularities of the English language.  

\section*{Acknowledgment}

The other authors thank Robert Vacareanu for providing the idea for this research and contributing with valuable insights in the development process.


The first author was partially supported by a private scholarship grant with id 31638/04.10.2023, given by the company Electrolux. 


\section*{Ethics}

We have extensively utilized off-the-self LLMs and NLI models, which may contain hidden biases. However, our approach makes inference more explicit, so these potential biases are more likely to be exposed during the morphing process.

We mostly used closed models, however our costs are reduced. We believe that this study does not exclude any communities from the point of view of the associated cost.

\bibliography{latex/acl_latex}

\begin{thebibliography}{48}
\providecommand{\natexlab}[1]{#1}

\bibitem[{Akoju et~al.(2023)Akoju, Vacareanu, Blanco, Riaz, and Surdeanu}]{akoju-etal-2023-synthetic}
Sushma~Anand Akoju, Robert Vacareanu, Eduardo Blanco, Haris Riaz, and Mihai Surdeanu. 2023.
\newblock \href {https://doi.org/10.18653/v1/2023.nlrse-1.12} {Synthetic dataset for evaluating complex compositional knowledge for natural language inference}.
\newblock In \emph{Proceedings of the 1st Workshop on Natural Language Reasoning and Structured Explanations (NLRSE)}, pages 157--168, Toronto, Canada. Association for Computational Linguistics.

\bibitem[{Angeli and Manning(2014)}]{angeli-manning-2014-naturalli}
Gabor Angeli and Christopher~D. Manning. 2014.
\newblock \href {https://doi.org/10.3115/v1/D14-1059} {{N}atural{LI}: Natural logic inference for common sense reasoning}.
\newblock In \emph{Proceedings of the 2014 Conference on Empirical Methods in Natural Language Processing ({EMNLP})}, pages 534--545, Doha, Qatar. Association for Computational Linguistics.

\bibitem[{Awasthi et~al.(2019)Awasthi, Sarawagi, Goyal, Ghosh, and Piratla}]{awasthi-etal-2019-parallel}
Abhijeet Awasthi, Sunita Sarawagi, Rasna Goyal, Sabyasachi Ghosh, and Vihari Piratla. 2019.
\newblock \href {https://doi.org/10.18653/v1/D19-1435} {Parallel iterative edit models for local sequence transduction}.
\newblock In \emph{Proceedings of the 2019 Conference on Empirical Methods in Natural Language Processing and the 9th International Joint Conference on Natural Language Processing (EMNLP-IJCNLP)}, pages 4260--4270, Hong Kong, China. Association for Computational Linguistics.

\bibitem[{Bowman et~al.(2015)Bowman, Angeli, Potts, and Manning}]{bowman2015large}
Samuel~R Bowman, Gabor Angeli, Christopher Potts, and Christopher~D Manning. 2015.
\newblock A large annotated corpus for learning natural language inference.
\newblock \emph{arXiv preprint arXiv:1508.05326}.

\bibitem[{Camburu et~al.(2018)Camburu, Rockt\"{a}schel, Lukasiewicz, and Blunsom}]{camburu2018explainablenli}
Oana-Maria Camburu, Tim Rockt\"{a}schel, Thomas Lukasiewicz, and Phil Blunsom. 2018.
\newblock \href {https://proceedings.neurips.cc/paper_files/paper/2018/file/4c7a167bb329bd92580a99ce422d6fa6-Paper.pdf} {e-snli: Natural language inference with natural language explanations}.
\newblock In \emph{Advances in Neural Information Processing Systems}, volume~31. Curran Associates, Inc.

\bibitem[{Camburu et~al.(2020)Camburu, Shillingford, Minervini, Lukasiewicz, and Blunsom}]{camburu-etal-2020-make}
Oana-Maria Camburu, Brendan Shillingford, Pasquale Minervini, Thomas Lukasiewicz, and Phil Blunsom. 2020.
\newblock \href {https://doi.org/10.18653/v1/2020.acl-main.382} {Make up your mind! adversarial generation of inconsistent natural language explanations}.
\newblock In \emph{Proceedings of the 58th Annual Meeting of the Association for Computational Linguistics}, pages 4157--4165, Online. Association for Computational Linguistics.

\bibitem[{Condoravdi et~al.(2003)Condoravdi, Crouch, de~Paiva, Stolle, and Bobrow}]{condoravdi-etal-2003-entailment}
Cleo Condoravdi, Dick Crouch, Valeria de~Paiva, Reinhard Stolle, and Daniel~G. Bobrow. 2003.
\newblock \href {https://aclanthology.org/W03-0906} {Entailment, intensionality and text understanding}.
\newblock In \emph{Proceedings of the {HLT}-{NAACL} 2003 Workshop on Text Meaning}, pages 38--45.

\bibitem[{Dagan et~al.(2005)Dagan, Glickman, and Magnini}]{Dagan2005ThePR}
Ido Dagan, Oren Glickman, and Bernardo Magnini. 2005.
\newblock \href {https://api.semanticscholar.org/CorpusID:8587959} {The pascal recognising textual entailment challenge}.
\newblock In \emph{Machine Learning Challenges Workshop}.

\bibitem[{Devlin et~al.(2019)Devlin, Chang, Lee, and Toutanova}]{Devlin2019BERTPO}
Jacob Devlin, Ming-Wei Chang, Kenton Lee, and Kristina Toutanova. 2019.
\newblock \href {https://api.semanticscholar.org/CorpusID:52967399} {Bert: Pre-training of deep bidirectional transformers for language understanding}.
\newblock In \emph{North American Chapter of the Association for Computational Linguistics}.

\bibitem[{Feng et~al.(2022)Feng, Yang, Zhu, and Greenspan}]{feng-etal-2022-neuro}
Yufei Feng, Xiaoyu Yang, Xiaodan Zhu, and Michael Greenspan. 2022.
\newblock \href {https://doi.org/10.1162/tacl_a_00458} {Neuro-symbolic natural logic with introspective revision for natural language inference}.
\newblock \emph{Transactions of the Association for Computational Linguistics}, 10:240--256.

\bibitem[{Golchin and Surdeanu(2024)}]{golchin2024time}
Shahriar Golchin and Mihai Surdeanu. 2024.
\newblock \href {https://openreview.net/forum?id=2Rwq6c3tvr} {Time travel in {LLM}s: Tracing data contamination in large language models}.
\newblock In \emph{Proceedings of the Twelfth International Conference on Learning Representations (ICLR)}.

\bibitem[{Guu et~al.(2018)Guu, Hashimoto, Oren, and Liang}]{guu-etal-2018-generating}
Kelvin Guu, Tatsunori~B. Hashimoto, Yonatan Oren, and Percy Liang. 2018.
\newblock \href {https://doi.org/10.1162/tacl_a_00030} {Generating sentences by editing prototypes}.
\newblock \emph{Transactions of the Association for Computational Linguistics}, 6:437--450.

\bibitem[{Hu and Moss(2018)}]{hu-moss-2018-polarity}
Hai Hu and Larry Moss. 2018.
\newblock \href {https://doi.org/10.18653/v1/S18-2015} {Polarity computations in flexible categorial grammar}.
\newblock In \emph{Proceedings of the Seventh Joint Conference on Lexical and Computational Semantics}, pages 124--129, New Orleans, Louisiana. Association for Computational Linguistics.

\bibitem[{Huang et~al.(2018)Huang, Wu, Wei, and Zhou}]{huang2018text}
Shaohan Huang, Yu~Wu, Furu Wei, and Ming Zhou. 2018.
\newblock Text morphing.
\newblock \emph{arXiv preprint arXiv:1810.00341}.

\bibitem[{Jiang et~al.(2019)Jiang, He, Chen, Liu, Gao, and Zhao}]{Jiang2019SMARTRA}
Haoming Jiang, Pengcheng He, Weizhu Chen, Xiaodong Liu, Jianfeng Gao, and Tuo Zhao. 2019.
\newblock \href {https://api.semanticscholar.org/CorpusID:207847598} {Smart: Robust and efficient fine-tuning for pre-trained natural language models through principled regularized optimization}.
\newblock In \emph{Annual Meeting of the Association for Computational Linguistics}.

\bibitem[{Jiang and de~Marneffe(2022)}]{jiang-marneffe-2022-investigating}
Nan-Jiang Jiang and Marie-Catherine de~Marneffe. 2022.
\newblock \href {https://doi.org/10.1162/tacl_a_00523} {Investigating reasons for disagreement in natural language inference}.
\newblock \emph{Transactions of the Association for Computational Linguistics}, 10:1357--1374.

\bibitem[{Korakakis and Vlachos(2023)}]{korakakis-vlachos-2023-improving}
Michalis Korakakis and Andreas Vlachos. 2023.
\newblock \href {https://doi.org/10.18653/v1/2023.acl-long.801} {Improving the robustness of {NLI} models with minimax training}.
\newblock In \emph{Proceedings of the 61st Annual Meeting of the Association for Computational Linguistics (Volume 1: Long Papers)}, pages 14322--14339, Toronto, Canada. Association for Computational Linguistics.

\bibitem[{Krishna et~al.(2021)Krishna, Riedel, and Vlachos}]{Krishna2021ProoFVerNL}
Amrith Krishna, Sebastian Riedel, and Andreas Vlachos. 2021.
\newblock \href {https://api.semanticscholar.org/CorpusID:237289760} {Proofver: Natural logic theorem proving for fact verification}.
\newblock \emph{Transactions of the Association for Computational Linguistics}, 10:1013--1030.

\bibitem[{Krishna et~al.(2022)Krishna, Riedel, and Vlachos}]{krishna2022proofver}
Amrith Krishna, Sebastian Riedel, and Andreas Vlachos. 2022.
\newblock Proofver: Natural logic theorem proving for fact verification.
\newblock \emph{Transactions of the Association for Computational Linguistics}, 10:1013--1030.

\bibitem[{Kumar and Talukdar(2020)}]{Kumar2020NILEN}
Sawan Kumar and Partha~Pratim Talukdar. 2020.
\newblock \href {https://api.semanticscholar.org/CorpusID:218869840} {Nile : Natural language inference with faithful natural language explanations}.
\newblock In \emph{Annual Meeting of the Association for Computational Linguistics}.

\bibitem[{Lakoff(1970)}]{Lakoff1970LinguisticsAN}
George Lakoff. 1970.
\newblock \href {https://api.semanticscholar.org/CorpusID:46974778} {Linguistics and natural logic}.
\newblock \emph{Synthese}, 22:151--271.

\bibitem[{Lewis et~al.(2019)Lewis, Liu, Goyal, Ghazvininejad, rahman Mohamed, Levy, Stoyanov, and Zettlemoyer}]{Lewis2019BARTDS}
Mike Lewis, Yinhan Liu, Naman Goyal, Marjan Ghazvininejad, Abdel rahman Mohamed, Omer Levy, Veselin Stoyanov, and Luke Zettlemoyer. 2019.
\newblock \href {https://api.semanticscholar.org/CorpusID:204960716} {Bart: Denoising sequence-to-sequence pre-training for natural language generation, translation, and comprehension}.
\newblock In \emph{Annual Meeting of the Association for Computational Linguistics}.

\bibitem[{Liu et~al.(2019)Liu, Ott, Goyal, Du, Joshi, Chen, Levy, Lewis, Zettlemoyer, and Stoyanov}]{Liu2019RoBERTaAR}
Yinhan Liu, Myle Ott, Naman Goyal, Jingfei Du, Mandar Joshi, Danqi Chen, Omer Levy, Mike Lewis, Luke Zettlemoyer, and Veselin Stoyanov. 2019.
\newblock \href {https://api.semanticscholar.org/CorpusID:198953378} {Roberta: A robustly optimized bert pretraining approach}.
\newblock \emph{ArXiv}, abs/1907.11692.

\bibitem[{MacCartney and Manning(2007)}]{MacCartney2007NaturalLF}
Bill MacCartney and Christopher~D. Manning. 2007.
\newblock \href {https://api.semanticscholar.org/CorpusID:9925526} {Natural logic for textual inference}.
\newblock In \emph{ACL-PASCAL@ACL}.

\bibitem[{MacCartney and Manning(2008)}]{MacCartney2008ModelingSC}
Bill MacCartney and Christopher~D. Manning. 2008.
\newblock \href {https://api.semanticscholar.org/CorpusID:5617715} {Modeling semantic containment and exclusion in natural language inference}.
\newblock In \emph{International Conference on Computational Linguistics}.

\bibitem[{MacCartney and Manning(2009{\natexlab{a}})}]{maccartney2009extended}
Bill MacCartney and Christopher~D Manning. 2009{\natexlab{a}}.
\newblock An extended model of natural logic.
\newblock In \emph{Proceedings of the eight international conference on computational semantics}, pages 140--156.

\bibitem[{MacCartney and Manning(2009{\natexlab{b}})}]{MacCartney2009AnEM}
Bill MacCartney and Christopher~D. Manning. 2009{\natexlab{b}}.
\newblock \href {https://api.semanticscholar.org/CorpusID:6561519} {An extended model of natural logic}.
\newblock In \emph{International Conference on Computational Semantics}.

\bibitem[{MacCartney and Manning(2014)}]{maccartney2014natural}
Bill MacCartney and Christopher~D Manning. 2014.
\newblock Natural logic and natural language inference.
\newblock In \emph{Computing Meaning: Volume 4}, pages 129--147. Springer.

\bibitem[{Marelli et~al.(2014)Marelli, Menini, Baroni, Bentivogli, Bernardi, and Zamparelli}]{sick}
Marco Marelli, Stefano Menini, Marco Baroni, Luisa Bentivogli, Raffaella Bernardi, and Roberto Zamparelli. 2014.
\newblock A {SICK} cure for the evaluation of compositional distributional semantic models.
\newblock In \emph{Proceedings of the Ninth International Conference on Language Resources and Evaluation (LREC’14)}, pages 216–--223.

\bibitem[{Raffel et~al.(2019)Raffel, Shazeer, Roberts, Lee, Narang, Matena, Zhou, Li, and Liu}]{Raffel2019ExploringTL}
Colin Raffel, Noam~M. Shazeer, Adam Roberts, Katherine Lee, Sharan Narang, Michael Matena, Yanqi Zhou, Wei Li, and Peter~J. Liu. 2019.
\newblock \href {https://api.semanticscholar.org/CorpusID:204838007} {Exploring the limits of transfer learning with a unified text-to-text transformer}.
\newblock \emph{J. Mach. Learn. Res.}, 21:140:1--140:67.

\bibitem[{Reid et~al.(2023)Reid, Hellendoorn, and Neubig}]{reid2023diffuser}
Machel Reid, Vincent~Josua Hellendoorn, and Graham Neubig. 2023.
\newblock \href {https://openreview.net/forum?id=nG9RF9z1yy3} {Diffus{ER}: Diffusion via edit-based reconstruction}.
\newblock In \emph{The Eleventh International Conference on Learning Representations}.

\bibitem[{Reid and Neubig(2022)}]{reid-neubig-2022-learning}
Machel Reid and Graham Neubig. 2022.
\newblock \href {https://doi.org/10.18653/v1/2022.findings-emnlp.280} {Learning to model editing processes}.
\newblock In \emph{Findings of the Association for Computational Linguistics: EMNLP 2022}, pages 3822--3832, Abu Dhabi, United Arab Emirates. Association for Computational Linguistics.

\bibitem[{Reimers and Gurevych(2019)}]{reimers-2019-sentence-bert}
Nils Reimers and Iryna Gurevych. 2019.
\newblock \href {https://arxiv.org/abs/1908.10084} {Sentence-bert: Sentence embeddings using siamese bert-networks}.
\newblock In \emph{Proceedings of the 2019 Conference on Empirical Methods in Natural Language Processing}. Association for Computational Linguistics.

\bibitem[{Ribeiro et~al.(2016)Ribeiro, Singh, and Guestrin}]{ribeiro2016whyitrustyou}
Marco~Tulio Ribeiro, Sameer Singh, and Carlos Guestrin. 2016.
\newblock \href {https://arxiv.org/abs/1602.04938} {"why should i trust you?": Explaining the predictions of any classifier}.
\newblock \emph{Preprint}, arXiv:1602.04938.

\bibitem[{Rozanova et~al.(2022)Rozanova, Ferreira, Thayaparan, Valentino, and Freitas}]{rozanova-etal-2022-decomposing}
Julia Rozanova, Deborah Ferreira, Mokanarangan Thayaparan, Marco Valentino, and Andre Freitas. 2022.
\newblock \href {https://doi.org/10.18653/v1/2022.blackboxnlp-1.33} {Decomposing natural logic inferences for neural {NLI}}.
\newblock In \emph{Proceedings of the Fifth BlackboxNLP Workshop on Analyzing and Interpreting Neural Networks for NLP}, pages 394--403, Abu Dhabi, United Arab Emirates (Hybrid). Association for Computational Linguistics.

\bibitem[{Sainz et~al.(2024)Sainz, Campos, García-Ferrero, Etxaniz, and Agirre}]{contindex}
Oscar Sainz, Jon~Ander Campos, Iker García-Ferrero, Julen Etxaniz, and Eneko Agirre. 2024.
\newblock \href {https://hitz-zentroa.github.io/lm-contamination/} {Llm contamination index}.
\newblock Last accessed: October 11, 2024.

\bibitem[{Sainz et~al.(2022)Sainz, Gonzalez-Dios, Lopez~de Lacalle, Min, and Agirre}]{sainz-etal-2022-textual}
Oscar Sainz, Itziar Gonzalez-Dios, Oier Lopez~de Lacalle, Bonan Min, and Eneko Agirre. 2022.
\newblock \href {https://doi.org/10.18653/v1/2022.findings-naacl.187} {Textual entailment for event argument extraction: Zero- and few-shot with multi-source learning}.
\newblock In \emph{Findings of the Association for Computational Linguistics: NAACL 2022}, pages 2439--2455, Seattle, United States. Association for Computational Linguistics.

\bibitem[{Sainz et~al.(2021)Sainz, Lopez~de Lacalle, Labaka, Barrena, and Agirre}]{sainz-etal-2021-label}
Oscar Sainz, Oier Lopez~de Lacalle, Gorka Labaka, Ander Barrena, and Eneko Agirre. 2021.
\newblock \href {https://doi.org/10.18653/v1/2021.emnlp-main.92} {Label verbalization and entailment for effective zero and few-shot relation extraction}.
\newblock In \emph{Proceedings of the 2021 Conference on Empirical Methods in Natural Language Processing}, pages 1199--1212, Online and Punta Cana, Dominican Republic. Association for Computational Linguistics.

\bibitem[{Storks et~al.(2019)Storks, Gao, and Chai}]{Storks2019RecentAI}
Shane Storks, Qiaozi Gao, and Joyce~Yue Chai. 2019.
\newblock \href {https://api.semanticscholar.org/CorpusID:213613608} {Recent advances in natural language inference: A survey of benchmarks, resources, and approaches}.
\newblock \emph{arXiv: Computation and Language}.

\bibitem[{Strong et~al.(2024)Strong, Aly, and Vlachos}]{Strong2024ZeroShotFV}
Marek Strong, Rami Aly, and Andreas Vlachos. 2024.
\newblock \href {https://api.semanticscholar.org/CorpusID:273162463} {Zero-shot fact verification via natural logic and large language models}.

\bibitem[{Sun et~al.(2020)Sun, Fan, Han, Sun, Meng, Wu, and Li}]{Sun2020SelfExplainingSI}
Zijun Sun, Chun Fan, Qinghong Han, Xiaofei Sun, Yuxian Meng, Fei Wu, and Jiwei Li. 2020.
\newblock \href {https://api.semanticscholar.org/CorpusID:227254768} {Self-explaining structures improve nlp models}.
\newblock \emph{ArXiv}, abs/2012.01786.

\bibitem[{Thorne et~al.(2019)Thorne, Vlachos, Christodoulopoulos, and Mittal}]{Thorne2019GeneratingTE}
James Thorne, Andreas Vlachos, Christos Christodoulopoulos, and Arpit Mittal. 2019.
\newblock \href {https://api.semanticscholar.org/CorpusID:129945615} {Generating token-level explanations for natural language inference}.
\newblock \emph{ArXiv}, abs/1904.10717.

\bibitem[{Turpin et~al.(2024)Turpin, Michael, Perez, and Bowman}]{turpin2024language}
Miles Turpin, Julian Michael, Ethan Perez, and Samuel Bowman. 2024.
\newblock Language models don't always say what they think: unfaithful explanations in chain-of-thought prompting.
\newblock \emph{Advances in Neural Information Processing Systems}, 36.

\bibitem[{Vaswani et~al.(2017)Vaswani, Shazeer, Parmar, Uszkoreit, Jones, Gomez, Kaiser, and Polosukhin}]{Vaswani2017AttentionIA}
Ashish Vaswani, Noam~M. Shazeer, Niki Parmar, Jakob Uszkoreit, Llion Jones, Aidan~N. Gomez, Lukasz Kaiser, and Illia Polosukhin. 2017.
\newblock \href {https://api.semanticscholar.org/CorpusID:13756489} {Attention is all you need}.
\newblock In \emph{Neural Information Processing Systems}.

\bibitem[{Wang et~al.(2021)Wang, Fang, Khabsa, Mao, and Ma}]{Wang2021EntailmentAF}
Sinong Wang, Han Fang, Madian Khabsa, Hanzi Mao, and Hao Ma. 2021.
\newblock Entailment as few-shot learner.
\newblock \emph{ArXiv}, abs/2104.14690.

\bibitem[{Wei et~al.(2024)Wei, Wang, Schuurmans, Bosma, Ichter, Xia, Chi, Le, and Zhou}]{Chain-of-Thought}
Jason Wei, Xuezhi Wang, Dale Schuurmans, Maarten Bosma, Brian Ichter, Fei Xia, Ed~H. Chi, Quoc~V. Le, and Denny Zhou. 2024.
\newblock Chain-of-thought prompting elicits reasoning in large language models.
\newblock In \emph{Proceedings of the 36th International Conference on Neural Information Processing Systems}, NIPS '22. Curran Associates Inc.

\bibitem[{Williams et~al.(2018)Williams, Nangia, and Bowman}]{williams-etal-2018-broad}
Adina Williams, Nikita Nangia, and Samuel Bowman. 2018.
\newblock \href {https://doi.org/10.18653/v1/N18-1101} {A broad-coverage challenge corpus for sentence understanding through inference}.
\newblock In \emph{Proceedings of the 2018 Conference of the North {A}merican Chapter of the Association for Computational Linguistics: Human Language Technologies, Volume 1 (Long Papers)}, pages 1112--1122, New Orleans, Louisiana. Association for Computational Linguistics.

\bibitem[{Yin et~al.(2019)Yin, Hay, and Roth}]{yin-etal-2019-benchmarking}
Wenpeng Yin, Jamaal Hay, and Dan Roth. 2019.
\newblock \href {https://doi.org/10.18653/v1/D19-1404} {Benchmarking zero-shot text classification: Datasets, evaluation and entailment approach}.
\newblock In \emph{Proceedings of the 2019 Conference on Empirical Methods in Natural Language Processing and the 9th International Joint Conference on Natural Language Processing (EMNLP-IJCNLP)}, pages 3914--3923, Hong Kong, China. Association for Computational Linguistics.

\end{thebibliography}
\newpage
\appendix
%
%

\section{Morphism generation examples}
\label{sec:appendix_examples}

Figure \ref{fig:prompt_example} presents the prompt that we have used for the generation of the synthetic dataset labeled with morphisms. The first part of the prompt consists on giving basic rules on the morphing task such as specifying what is the input and the desired output, the maximum number of intermediary sentences and the operations used (with their structure). The next part presents in more detail information about each operation and how they should be performed. This part was developed through prompt engineering, analyzing the systematic mistakes that the model was making in the generation process. Then, we give 12 examples of morphisms, together with the morph operations. Finally, the premise and hypothesis are given. Figure \ref{fig:ICL_example} presents a humanly annotated example in the prompt. The output structure of the LLM follows the structure of the ICL example. An example of morphism generated by the student model (fine-tuned GPT-4o-mini) is presented in Figure \ref{fig:output_example}.

Figure \ref{fig:explanation_prompt} shows the prompt used for generating the GPT-4o and Llama 3.1 8B explanations that were compared against \name\ explanations.

\begin{figure*}
\begin{center}
\resizebox{1\textwidth
}{!}{
\begin{tabular}{p{16cm}}
\hline
Take a deep breath and work on this problem step-by-step. Please generate intermediate sentences from `Sentence 1` to `Sentence 2`, essentially morphing `Sentence 1` to `Sentence 2` through successive atomic edits. Each edit gives another interpolated sentence. Limit the number of interpolation/changes to at most 7. The atomic edits that you are allowed to do have the following structure, manipulating short parts of text:\\ \\
        1. Replace operations - (replace, <old\_text>, <new\_text>)\\
	2. Remove operations - (remove, <text>)\\
        3. Insert operations - (insert, <text>) \\\\
            
You are required to do all the operations in the order specified above: first just replacements, then removals and lastly insertions if needed. Each edit must consider similar syntactic groups, so you are not allowed to break syntactic boundaries. Perform multiple small operations, rather than one operation that changes the whole text. For example, a replace operation that changes most of the text could be broken down into multiple replace operations, followed by remove operations.\\
\\
The replacements are the most usual operations. These operations must be done by comparing syntactically similar word groups from the current sentence with ones from the target sentence (Sentence 2). The replacement text may be a rephrase of the starting text group, or even a contradictory statement, depending on the form and meaning of the final sentence. \\ \\
The removals are done on groups of text that are totally unrelated to the final sentence. This includes additional information specified in the starting sentence that disappears while morphing towards the final sentence.\\ \\
The insertions represent new text that is unrelated to the knowledge presented in the starting sentence. Do inserts only when it is necessary, when inserting text that is totally unrelated to the textual constructs of the current sentence. Under no circumstances you are allowed to remove certain words then insert related words. This should be done using a replacement operation instead.\\ \\
I will give some examples below. Keep the same structure of your response as seen in the examples, with no additional text/explanations.\\ \\

\{Example\ 1\}\\
\{Example 2\}\\
... \\
\{Example 12\}\\ \\
Generate the intermediate sentences and print the atomic edits for the following pair of sentences:\\

Sentence 1:\\
\{Premise\}\\

Sentence 2:\\
\{Hypothesis\}\\ \\

Morphism:\\
\hline
\end{tabular}
}
\end{center}
\caption{Prompt example for morphism generation.}
\label{fig:prompt_example}
\end{figure*}

\begin{figure}[!h]
\begin{center}
\resizebox{1\linewidth
}{!}{
\begin{tabular}{p{8cm}}
\hline
Sentence 1: \\
A man with a white beard speaks into a microphone wearing a long-sleeved gray button down shirt. \\ 
Sentence 2: \\
A man with a white beard is sitting quietly on a couch.\\ \\
Morphism:\\\\

-Replacements:\\
(replace, speaks into a microphone, is sitting quietly)\\ 
A man with a white beard is sitting quietly wearing a long-sleeved gray button down shirt. \\ \\
-Removals:\\
(remove, wearing a long-sleeved gray button down shirt) \\ 
A man with a white beard is sitting quietly.  \\ \\
-Insertions:\\
(insert, on a couch) \\ 
A man with a white beard is sitting quietly on a couch.\\

\hline
\end{tabular}
}
\end{center}
\vspace{-1mm}
\caption{Example of a manually annotated morphism from the ICL pool.}
\label{fig:ICL_example}
\end{figure}

\begin{figure}[!h]
\begin{center}
\resizebox{1\linewidth
}{!}{
\begin{tabular}{p{8cm}}
\hline

Morphism:\\\\

-Replacements:\\
(replace, A white man is walking a dog, A dog with a brown and white coat is trotting) \\
A dog with a brown and white coat is trotting through brown water with difficulty.\\ 
(replace, brown water, shallow water)\\
A dog with a brown and white coat is trotting through shallow water with difficulty.\\ \\
-Removals:\\
(remove, with difficulty)\\
A dog with a brown and white coat is trotting through shallow water.\\ \\
-Insertions:\\

\hline
\end{tabular}
}
\end{center}
\vspace{-1mm}
\caption{Example of a morphism as in the LLM output. The premise is "A white man is walking a dog through brown water with difficulty".}
\label{fig:output_example}
\end{figure}

\begin{figure}[!h]
\begin{center}
\resizebox{1\linewidth
}{!}{
\begin{tabular}{p{8cm}}
\hline
You have to provide the label and explanations for a Natural Language Inference (NLI) task. Natural Language Inference is the task of determining whether a "hypothesis" is true (entailment), false (contradiction), or undetermined (neutral) given a "premise".
You will be given the premise and the hypothesis, and must state if they have an entailment, contradiction or neutral relation. You are then required to provide the reasoning process that explains why the label applies for the pair of sentences.
The explanations must be clear and concise, using natural language.   \\ \\
Premise: \\
{Premise} \\ \\
Hypothesis:\\
{Hypothesis} \\ \\
Label and Reasoning process:\\

\hline
\end{tabular}
}
\end{center}
\vspace{-1mm}
\caption{The prompt used for generating the GPT-4o and Llama 3.1 8B explanations.}
\label{fig:explanation_prompt}
\end{figure}

\section{Results on validation datasets}
\label{sec:appendix_dev_dataset}

Tables \ref{tab:SICK_val} and \ref{tab:MNLI_val} present the results on the validation datasets of SICK and MNLI. We observe the same behaviour as the one described in Section \ref{sec:results} on the test datasets. Our method significantly outperforms the state-of-the-art models in the OOD setting, with an increase of 1.21\% for RoBERTa and 4.84\% for BART in the case of SICK, and 3.10\% for RoBERTa and 4.59\% for BART in the case of MNLI without voice normalization. 

\begin{table}[!h]
\begin{center}
\resizebox{1\linewidth
}{!}{
\begin{tabularx}{1.2\linewidth}{ l *{2}{Z} }
\hline
\textbf{SICK} & \textbf{ID}& \textbf{OOD} \\
\hline
RoBERTa Vanilla  & 89.90 & 57.78 \\
RoBERTa Vanilla (+VN) & \textbf{90.91} & 58.38 \\
RoBERTa Morphism (+VN) & 85.86 & \textbf{58.99}\\
\hline
BART Vanilla  & 89.70 & 59.60   \\
BART Vanilla (+VN) & \textbf{89.90} & 61.01   \\
BART Morphism (+VN) &  87.68 & \textbf{64.44}   \\
\hline
\end{tabularx}
}
\end{center}
\vspace{-1mm}
\caption{\name\ accuracy on the SICK validation dataset, using two NLI engines: RoBERTa and BART. We compare our results against the two ``vanilla'' NLI models, i.e., without using text morphing. For OOD, we use the RoBERTa models trained on MNLI, and BART models trained on SNLI, MNLI and FEVER.}
\label{tab:SICK_val}
\end{table}

\begin{table}[!h]
\begin{center}
\resizebox{1\linewidth
}{!}{
\begin{tabularx}{1.2\linewidth}{ l *{2}{Z} }
\hline
\textbf{MNLI} & \textbf{ID}& \textbf{OOD} \\
\hline
RoBERTa Vanilla  & \textbf{90.12} & 55.43 \\
RoBERTa Vanilla (+VN) & 88.59 & 54.61 \\
RoBERTa Morphism  & 84.71 & \textbf{58.53} \\
RoBERTa Morphism (+VN) & 83.13 & 57.94 \\
\hline
BART Vanilla  & \textbf{89.62} & 48.67   \\
BART Vanilla (+VN) & 87.70 & 48.25  \\
BART Morphism  &  83.66 & \textbf{53.26} \\
BART Morphism (+VN) &  81.70 & 52.65   \\
\hline
\end{tabularx}
}
\end{center}
\vspace{-1mm}
\caption{\name\ accuracy on the MNLI validation dataset, under the same settings as Table \ref{tab:SICK_val}. For the OOD results, we train the respective NLI model on SICK.}
\label{tab:MNLI_val}
\end{table}

\section{NLI models and LLMs used}
\label{sec:models_used}

Throughout our study, we used NLI classifiers in order to generate the individual labels between each pair of sentences and to provide a comparison baseline. Table \ref{tab:nli_models} shows the NLI models used, together with the dataset they were fine-tuned on and their source (Hugging Face path\footnote{https://huggingface.co/}). As we could not find an off-the-shelf BART-large model fine-tuned on SICK, we fine-tuned a version of our own on the train split. We have used a learning rate of 1e-4 with 500 warm-up steps and batch size of 32 for 5 epochs. Cross Entropy is used as loss function, together with AdamW as optimizer. 

\begin{table*}[!ht]
\begin{center}
\resizebox{1\textwidth
}{!}{
\begin{tabular}{l l l}
\hline
 \textbf{Model type} & \textbf{Dataset} & \textbf{Huggingface path} \\
\hline
RoBERTa-large&  SICK&  varun-v-rao/roberta-large-fp-sick \\
RoBERTa-large&  MNLI&  FacebookAI/roberta-large-mnli \\
BART-large&  SICK&  (fine-tuned in-house) \\
BART-large& SNLI+MNLI+FEVER& ynie/bart-large-snli\_mnli\_fever\_anli\_R1\_R2\_R3-nli \\

\hline
\end{tabular}
}
\end{center}
\caption{The NLI models used for classification. }
\label{tab:nli_models}
\end{table*}

As mentioned in the article, we used various LLMs of the GPT-4 family. Here, we provide the identifiers of these models to increase reproducibility:
\begin{itemize}
    \item GPT-4o (labeling, explanations): gpt-4o-2024-08-06
    \item GPT-4 (teacher model): gpt-4-0125-preview
    \item GPT-4o-mini (student model, voice normalization, labeling): gpt-4o-mini-2024-07-18
\end{itemize}

The total budget for synthetically annotating morphisms using ICL, fine-tuning the student model, making the ablation studies, and testing the performance of our model was approximately 350\$.

It is important to mention that our method does not rely on proprietary models (GPT family). At the beginning of our research, we experimented with GPT-4, Claude 3 and Llama 3.1, and chose GPT-4 as it performed slightly better in the morphing generation process.

To verify whether our overall results hold with an open-weight LLM, we conducted an experiment in which we took a small sample size from SICK (50 examples) and generated morphisms with both GPT-4o-mini and Llama-3.1-70b-Instruct (using in-context learning examples). Then we labeled the morphisms using a RoBERTa based NLI model for both in-domain and out-of-domain scenarios. We present the results in Table \ref{tab:llama}. We observe that the results are reasonably similar. That is, the GPT model outperforms Llama for the in-domain case, but Llama is superior for the out-of-domain case. This small experiment yields promising insights into the applicability of our method to LLMs from other families.

\begin{table}[!ht]
\begin{center}
\resizebox{1\linewidth
}{!}{
\begin{tabularx}{1.2\linewidth}{ l *{2}{Z} }
\hline
\textbf{SICK 50 examples} & \textbf{ID}& \textbf{OOD} \\
\hline
Vanilla  & \textbf{82.00} & 56.00 \\
MorphNLI GPT-4o-mini & 80.00 & 56.00 \\
MorphNLI Llama-3.1-70b-Instruct & 72.00 & \textbf{62.00} \\

\hline
\end{tabularx}
}
\end{center}
\vspace{-1mm}
\caption{\name\ accuracy on a small sample from SICK, using GPT-4o-mini or Llama-3.1-70b-Instruct for generating the morphisms. The NLI modules used are RoBERTa based. }
\label{tab:llama}
\end{table}

\section{Lexical sensitivity}
\label{sec:appendix_lexical}

\begin{figure*}[!h]
    \begin{subfigure}[b]{0.5\textwidth}
        \centering
        \includegraphics[width=1\linewidth]{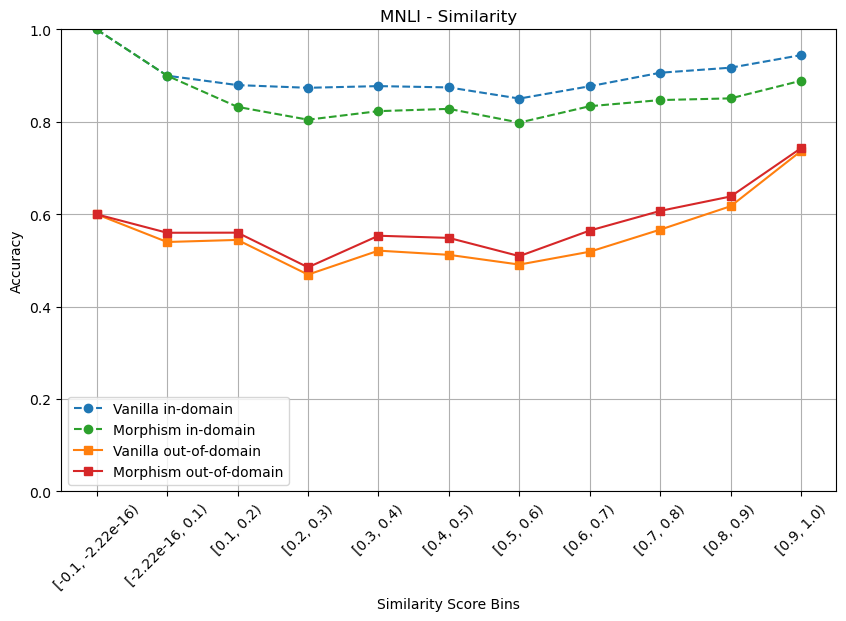}
        \subcaption{MNLI Similarity}
        \label{fig:MNLI_sim}
    \end{subfigure}
    \hfill
    \begin{subfigure}[b]{0.5\textwidth}
        \centering
        \includegraphics[width=1\linewidth]{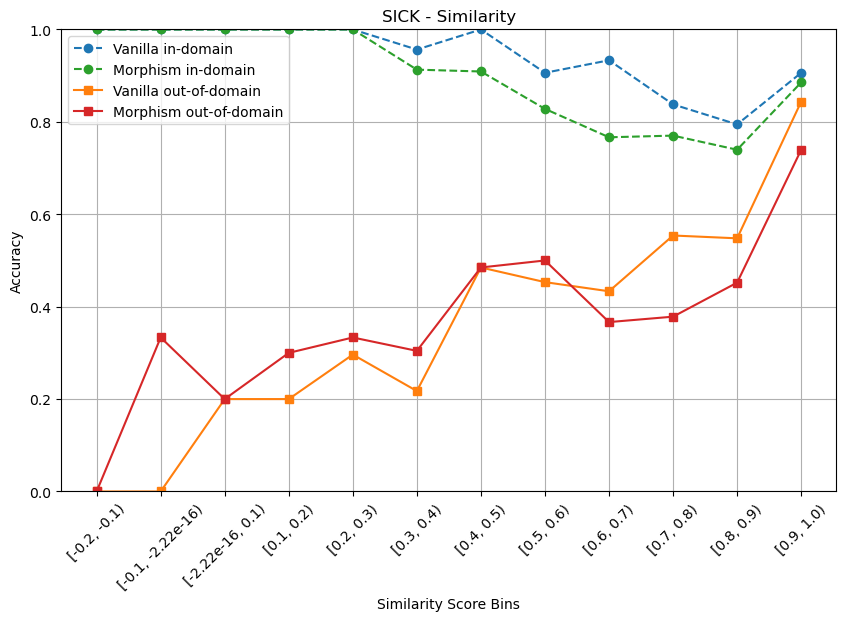}
        \subcaption{SICK Similarity}
        \label{fig:SICK_sim}
    \end{subfigure}
    \\
    \begin{subfigure}[b]{0.5\textwidth}
        \centering
        \includegraphics[width=1\linewidth]{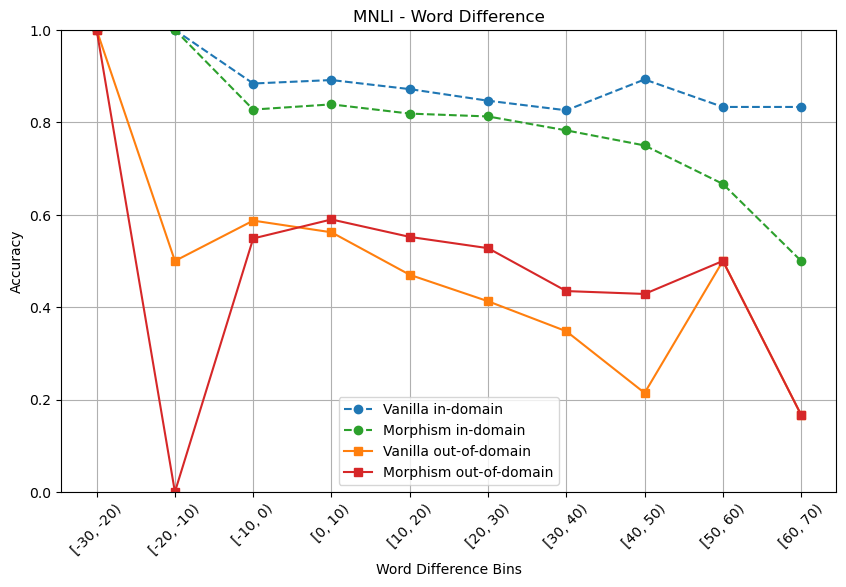}
        \subcaption{MNLI Word Difference}
        \label{fig:MNLI_wd}
    \end{subfigure}
    \hfill
    \begin{subfigure}[b]{0.5\textwidth}
        \centering
        \includegraphics[width=1\linewidth]{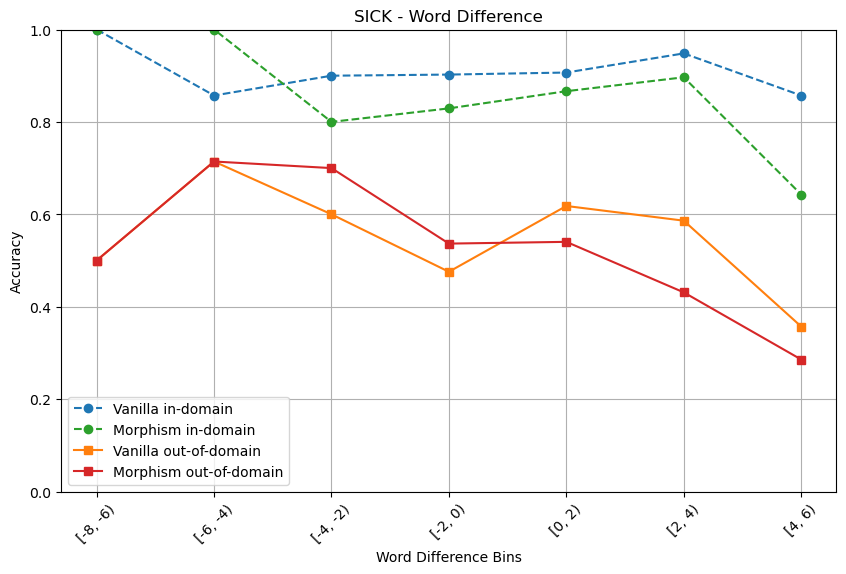}
        \subcaption{SICK Word Difference}
        \label{fig:SICK_wd}
    \end{subfigure}
    \caption{\name\ sensitivity to lexical difference in the premise and hypothesis pair. We can see that our model is less sensitive in the out-of-domain scenario and less prone to overfitting. }
    \label{fig:lexical}
    \vspace{-5mm}
\end{figure*}

We wanted to see how the performance of our approach varies considering the lexical difference between the premise and hypothesis. We measured the accuracy as the similarity between the hypothesis and the premise varies, and as the word difference varies. For the similarity, we used a sentence transformer (all-MiniLM-L6-v2) and measured the cosine similarity between premise and hypothesis. For the word difference, we computed the difference in words between the lemmatized premise and hypothesis. The results are presented in Figure \ref{fig:lexical}. We see that for both scenarios and datasets, \name\ is less sensitive to lexical differences, especially out-of-domain. We observe that in the case of word difference, the in-domain vanilla approach is not affected by a large difference. We consider this a clear sign of overfitting, especially as the rest of the methods have a drop in performance. This experiment further shows that our approach is less prone to overfitting and outperforms the vanilla models in out-of-domain scenarios. 

\end{document}